\documentclass[conference]{IEEEtran}
\IEEEoverridecommandlockouts
\usepackage{cite}
\usepackage{amsmath,amssymb,amsfonts}
\usepackage{graphicx}
\usepackage{textcomp}
\usepackage{xcolor}
\graphicspath{ {./images/} }
\usepackage[numbers]{natbib}
\setcitestyle{number}
\usepackage[shortlabels]{enumitem}

\usepackage[utf8]{inputenc} 
\usepackage[T1]{fontenc}    
\usepackage{hyperref}       
\usepackage{url}            
\usepackage{booktabs}       
\usepackage{amsfonts}       
\usepackage{nicefrac}       
\usepackage{microtype}      
\usepackage{algorithm}
\usepackage[noend]{algpseudocode}

\usepackage{color}

\usepackage{amsmath}
\usepackage[singlelinecheck=false]{caption}
\usepackage{placeins}

\newcommand{\rvec}[1]{\boldsymbol{\mathrm{{#1}}}}
\def\BibTeX{{\rm B\kern-.05em{\sc i\kern-.025em b}\kern-.08em
    T\kern-.1667em\lower.7ex\hbox{E}\kern-.125emX}}
\begin{document}

\title{MAIN: Multihead-Attention Imputation Networks\\
}


\author{
\IEEEauthorblockN{Spyridon Mouselinos\IEEEauthorrefmark{1}, 
Kyriakos Polymenakos\IEEEauthorrefmark{1}\IEEEauthorrefmark{2}, 
Antonis Nikitakis\IEEEauthorrefmark{1}, 
Konstantinos Kyriakopoulos\IEEEauthorrefmark{1}\IEEEauthorrefmark{3}}
\IEEEauthorblockA{\texttt{\{s.mouselinos, k.polymenakos, a.nikitakis, k.kyriakopoulos\}@deepsea.ai}}
\IEEEauthorblockA{\IEEEauthorrefmark{1}DeepSea Technologies, Athens, Greece}
\IEEEauthorblockA{\IEEEauthorrefmark{2} University of Oxford, Department of Engineering Science\\}
\IEEEauthorblockA{\IEEEauthorrefmark{3} University of Cambridge, Department of Engineering}}

\maketitle

\begin{abstract}
The problem of missing data, usually absent in curated and competition-standard datasets, is an unfortunate reality for most machine learning models used in industry applications. Recent work has focused on understanding the nature and the negative effects of such phenomena, while devising solutions for optimal imputation of the missing data, using both discriminative and generative approaches. 
We propose a novel mechanism based on multi-head attention which can be applied effortlessly in any model and achieves better downstream performance without the introduction of the full dataset in any part of the modeling pipeline.
Our method inductively models patterns of missingness in the input data in order to increase the performance of the downstream task. 
Finally, after evaluating our method against baselines for a number of datasets, we found performance gains that tend to be larger in scenarios of high missingness.

\end{abstract}


\section{Introduction}

Corrupted or missing data often raise the following questions:
If the missing values are imputed, what value is to be used for "padding"? 
What is the underlying relation with the normalization scheme of the input values? 
How should categorical inputs be treated? 
Popular approaches include replacing the missing values with their mean, in the case of continuous variables, or the median, in the case of discrete variables. 
Other approaches introduce a binary indicator of missingness that is jointly trained with the rest of the model parameters.

The simplest, although widely used, method is \emph{mean/mode impute}, where missing numerical values are replaced by their mean (0 in the Zeta Normalized Scale) and categorical values are replaced by their mode (the most frequent class of the dataset).
One disadvantage of mean/mode impute is that it can introduce ambiguity: variables that take values close to the mean or mode, might not be distinguished from variables whose missing values were imputed with the value of the mean or median.  
To resolve this ambiguity, a bitwise indicator (also referred to as \emph{mask}) was introduced, that assigns 1 to missing instances and 0 to non-missing instances. 
For simplicity, we refer to this method as Zero Impute (with) Mask Concat (\textbf{ZIMC}).

A straightforward extension to ZIMC is Sparsity Normalization (\textbf{SN}), 
which creates a per-example (horizontal) scaling by a constant factor \cite{vsp}, 
based on the global missingness factor. 
This method resolves the Variable Sparsity Problem, but fails to capture the dependencies in-between the features. 
In more detail, no adjustment is made based on what feature is missing and the values the existing features have.

Since the problem of missing data naturally defines a context of missing information we are inspired from feature-wise transformations \cite{fwt} as methods of context-based processing. More specifically the Feature Wise Linear Modulation (\textbf{FiLM}), naturally augments the idea of SN \cite{vsp} (\textbf{SN}) since it can be seen as a conditional scaling method. 
In order to efficiently capture the missing context we combine both the feature vector (i.e values) as well as the bitwise missingness indicator mask into contextual embeddings. 
In this regard, our method is capable of producing independent scaling coefficients for each individual feature, surpassing the \textbf{SN} method on various datasets and thus confirms that vertical (per-feature) conditioned imputation surpasses horizontal (per-instance) methods. Furthermore, we extend the attention mechanism of AimNet \cite{aimnet} into one that uses a richer key-value embedding representation, capable of encapsulating position, missingness and value information. 
On the contrary, in AimNet, only the missingness context is captured. 
That enables our model to continuously improve the embedding representations with both existing and missing features.

Our approach, Multihead-Attention Imputation Networks (\textbf{MAIN}), builds on \textbf{FiLM} as it creates multiple coefficients for each feature (multiple vertical imputation), based on the attention mechanism. 
In this modeling scenario, we create $N$ sets of $N$ coefficients where $N$ is the number of input features, with each set representing a conditioned modulation of a feature based on the value and existence of all features. 
This learnable combination mechanism captures both global feature-value and feature existence effects.

Briefly our method extends the SN method of \citet{vsp} in two ways:
\begin{enumerate}
\item learning a different weight coefficient for each individual feature - instead of the whole feature vector, thus dismissing the concept of linear covariation as proposed in SN\cite{vsp}. 
\item conditioning the coefficients not only on the existence vector of the input features but also on their observed values.
\end{enumerate}. Also our attention mechanism differs from \cite{aimnet} as we use multi-head attention and we build embeddings with both the value and the missing context of all the features.

\section{Related Work}


Imputation methods are often classified as \emph{discriminative} and \emph{generative}, based on the formulation of the missingness modeling mechanism.
In discriminative modeling, missing features are approximated directly through their conditional distributions, given the values of the existing features.
On the contrary, in generative modeling, missing features are often imputed in a two-step fashion: initially, the underlying joint distribution of all features is modeled, followed by a second step of conditionally generating the missing features from the existing information in the feature vector.

Discriminative methods depend on certain assumptions about the missingness mechanism.
Specifically, the data must conform to the Missing Completely At Random (MCAR) or - at a modest degree - Missing At Random (MAR) assumption \cite{missingness,misgan}. 
For MCAR, the probability of an entry missing is independent from the values of the other features, while for MAR, that probability depends on the values of the other features.
In both cases, the probability of a feature missing is assumed to be independent from the value of that feature.

Generative models focus on the joint probability distribution of the data instead of the conditional \cite{MIWAE,HI-VAE}. In \cite{MIWAE} the authors deal with problem of  missing data
with the use of deep latent variable models (DLVMs). Their approach is based on an importance-weighted autoencoder (IWAE) which maximises a potentially tight lower bound of the log-likelihood of the observed data. In \cite{HI-VAE} the authors  propose a general framework implementing VAEs to fit incomplete heterogenous data.
Generative models in general are more flexible and can effectively capture multimodal distributions, while discriminative models rely on point estimates.
The extra flexibility of the generative approach allows these models to deal with strongly MAR scenarios, as well as cases where the probability of a feature missing is dependent on the value of that feature (Missing Not At Random, or MNAR). On the other hand they are difficult to train and specifically, GANs can suffer from slow convergence and mode collapse problems.



It should be pointed that, joint modeling may excel in scenarios where the underlying connections between features are of spatial and/or auto-regressive nature (e.g image completion) \cite{inpaint,misgan,ambient}, but in case of tabular data this can be equally approached by (multiple) conditional modeling, since there is not such a variety of modalities that can only be approached by manifold walks.
 
Furthermore, when the final objective is to build a predictive task based on specific input (e.g regression, classification) the flexibility that generative modeling provides is unfit for time-critical applications.
In a scenario of new incoming information, a generative model will act in a "two-step fashion". Firstly, it will impute the missing data producing a complete feature vector and as a second step will apply an auxiliary or different architecture/model on the downstream task. Discriminative methods on the other hand can operate both in single and two-step fashion.
 
The work by \citet{vsp} gave an in-depth insight to the first two questions about data imputation, and re-introduced us with the phenomenon of Variable Sparsity Problem (VSP), while making significant progress towards the ``one-step'' approach. 
The VSP problem stresses an undesired phenomenon where the model's performance drops as its output significantly varies due to the rate of missingness in the given input also identified in \cite{dropblock}. 
The solution proposed by \citet{vsp}, named Sparsity Normalization (SN), seems to give a significant boost in the reconstruction capabilities of existing networks, while also increasing the robustness of a model's downstream task when dealing with missing inputs. 

In this work, we opt for a discriminative modeling schema, supporting the paradigm shift towards ``one-step'' methods, and study the effect of query-key-value attention performed in a multi-head fashion. We propose a custom key representation encoding, called Positional Encoded Vector (PEV). We show that even in heavily corrupted data our model incorporates the best attributes of discriminative and generative modeling, being an one-step method that is valid for both numeric and categorical inputs.
The contribution of this work can be summarized around 4 key ideas:
\begin{enumerate}
\item Due to its main application in real-time industrial scenarios, this work focuses on the performance gains of the downstream task(s), in contrast with other works that seem to sacrifice performance in favour of reconstruction of the full feature vector.

\item We show that our method has the best of both worlds between discriminative and generative methods as far as the downstream task performance is concerned.
It can handle both MCAR and MAR assumptions while not suffering from the limitation of training with complete vectors, that some discriminative and generative approaches silently imply. After testing its performance against 7 Datasets it shows consistent performance margin, as well as convergence without any instability issues.

\item We achieve a single-step, missingness-agnostic behaviour: our scheme can train on missing data directly, exploiting all the examples (including partially missing ones), and test on a different set never seen by the model, with completely untreated data (including missing features). This is not the case in autoencoder-based (i.e reconstruction) schemes like AimNet or generative methods \cite{gain}, where the model can train only on complete data where the missingness/reconstruction is artificially induced.
 
\item The PEV-based attention mechanism models both the missingness and the value of each individual feature in relation to the other features on the raw data distribution. AimNet \cite{aimnet} models only the missing context in a missing-induced distribution, while in VSP \cite{vsp} the feature vector is regularized horizontally, not taking into account each feature's individual variation.
\end{enumerate}

\section{Problem Formulation}
\subsection{Preliminaries}
Following the formulation of VSP \cite{vsp}, let $\rvec{x} \in \mathbb{R}^l$ and $\rvec{y} \in \mathbb{R}^d$ represent a training input/output pair instance of a model with $D$ tasks. Without loss of generality we will set $|D| = 1$ for the rest of our analysis.

To formulate missingness, we introduce $\rvec{m} \in \{0,1\}^l$, as the binary mask indicating missing values in $\rvec{x}$. 
Hence, we define $\rvec{x_{\text{miss}}}=\rvec{x} \odot \rvec{m}$, where $\odot$ denotes the Hadamard product (element-wise multiplication) of two vectors, as the corrupted input the model observes after the effect of the missingness mechanism. 
Finally, for the model let's assume a $N$-layer feed forward network, with convex, non-decreasing non-linearities $\sigma$ in each layer but the last. 
Each layer $i$ contains $n_i$ units, and we use $\rvec{W^i} \in \mathbb{R}^{{n_i} \times {n_{i-1}}}$ to denote the weight matrix, $\rvec{b^i} \in \mathbb{R}^{n_i}$ to denote the bias, and $\rvec{h^i} \in \mathbb{R}^{n_i}$ for the post-activation output vector. The calculation executed at each layer can then be written as $\rvec{h^i} = \sigma^i (\rvec{W^i} \rvec{h^{i-1}} + \rvec{b^i})$.\footnote{Throughout the paper we use uppercase and bold notation for matrices, lower and bold notation for vectors and lower and italic notation for elements. Superscripts without parentheses as $\rvec{h}^i$ are used to denote a quantity at layer $i$, and superscripts in parentheses, as $h^{(i)}$ to denote the $i$th element of  $\rvec{h}$.}

\subsection{Modeling Assumptions}
Our method works under the following two assumptions:
\begin{enumerate}
    \item Each element $m^k$ of the binary mask $\rvec{m}$ is MAR, meaning it may not depend on the other mask elements or the value of the element $x^k$, but it does depend on the values of the original input vector $x_i, i \neq k$, as described in \cite{missingness}. We will denote the means of the mask vector
    as $\mu_{m}$ and the means of the input vector as $\mu_{\text{missing}}$.
    
    \item The coordinates of $\rvec{b^i}$ and the elements of $\rvec{W^i}$ are mutually independent and follow the same distribution with means $\mu^{i}_{b}$ and $\mu^{i}_{w}$ respectively, as in \cite{vsp,he,bengio}. 
\end{enumerate}

\subsection{Methods}
\textbf{Theorem 1.} \textit{The expected value of the output layer of an N-Layer FFN with $l$-dimensional input and convex, non-decreasing, non-linearities under the MCAR assumption is bounded by}: $E[\rvec{h}^N] \geq f_N \circ f_{N-1} \circ \dots \circ f_1(\mu_{m}\mu_{x})$, where $f_{i}(x) = \sigma(\rvec{W}^{i} \cdot \rvec{x}^{i-1} + \rvec{b}^{i})$.

\vspace{3.5mm}
\textbf{Proof of 1.}
Proof is result of \cite{vsp}.

In Theorem 2 we extend the result of Theorem, for the more general MAR case.

\vspace{3.5mm}
\textbf{Theorem 2.} \textit{The expected value of the first layer of an N-Layer FFN of Theorem 1, under the MAR assumption, is bounded by: $E[\rvec{h}^1] \geq \sigma(n_0\mu^{1}_{w}\mu_{m}\mu_{\text{x}} + \mu^{1}_{b}) + T_1$, where $T_1 = \sigma(n_0\mu^{1}_{w}Cov(\rvec{x}^{1},\rvec{m}^{1}))$.}


\vspace{3.5mm}
\textbf{Proof of 2.}
For the output activation vector of the first layer $\rvec{h}^1$ it holds that:
\newline
\vspace{2mm}
$\rvec{h}^1 = \sigma(\rvec{W}^1 \times \rvec{x_{missing}} +\rvec{b}) = \sigma(\rvec{W}^1 \times \rvec{x} \odot \rvec{m} + \rvec{b})$. 
\newline
Then, by taking the expected value of both sides:
\newline
\vspace{2mm}
$E[\rvec{h}^1] = E[\sigma(\rvec{W}^1 \times \rvec{x} \odot \rvec{m} + \rvec{b})]$,
\newline
And due to non-decreasing convexity of $\sigma$:
\newline
$E[\rvec{h}^1] \geq \sigma(E[\rvec{W}^1 \times \rvec{x} \odot \rvec{m} + \rvec{b}])$. 

\vspace{2mm}
For the second part of the inequality it holds that:
\newline $\sigma(E[\rvec{W}^1 \times \rvec{x} \odot \rvec{m} + \rvec{b}])$ = $\sigma(E[\rvec{W}^1] \times E[\rvec{x} \odot \rvec{m}] + E[\rvec{b}]])$, based on assumption 2.

\vspace{2mm}
Focusing on elements of $E[\rvec{x} \odot \rvec{m}]$ we have that:
\begin{align*}
E[\rvec{x} \odot \rvec{m}] &= [E[x^{(1)} m^{(1)}], \dots, E[x^{(n_0)} m^{(n_0)}]]^T = \\
&=[\mu_m^{(1)} \mu_{\text{x}}^{(1)} + cov(x^{(1)}, m^{(1)}),\dots \\
&\dots,\mu_m^{(n_0)} \mu_{\text{x}}^{(n_0)} + cov(x^{(n_0)}, m^{(n_0)})]^T
\end{align*}

where $x^{(i)}, m^{(i)}$ refers to the $i$th component of the vector $\rvec{x}$ and 
$\rvec{m}$ respectively.

For the conditional probabilities $p(x_i | m_i)$, and $p(x_i)$ we have under:
\begin{align*}
\text{MCAR: }& p(x_i | m_i, x_1, \dots x_{i-1}, x_{i+1}, \dots, x_l) = \\
& p(x_i | m_i) = p(x_i) \\
\text{MAR: }& p(x_i |  x_1, \dots, x_{i-1}, x_{i+1}, \dots, x_l) = \\
&p(x_i | m_i, x_1, \dots, x_{i-1}, x_{i+1}, \dots, x_l)\\
&but\\
&p(x_i | m_i) \neq p(x_i)
\end{align*}

Recall that under the MCAR assumption  $cov(x^{(i)}, m^{(i)}) = 0$, as $\rvec{x},\rvec{m}$ are independent. 
Under the MAR assumption we are working with, $x^{(i)},m^{(i)}$ are conditionally independent, conditioned on the values of the rest of the features $x^{(j)}, i \neq j$, and in this case  $cov(x^{(i)}, m^{(i)}) $ is not necessarily $0$.

\vspace{2mm}
In that fashion the elementwise calculations lead to:
\begin{align*}
\sigma (E[&\rvec{W}^1] \times E[\rvec{x} \odot \rvec{m}]  + E[\rvec{b}]) = \sigma(E[\rvec{w}^{(1)}, \dots ,\rvec{w}^{(n_1)}] \\
& [\mu_m^{(1)} \mu_{\text{x}}^{(1)} +  cov(x^{(1)}, m^{(1)}),\dots, \mu_m^{(n_0)} \mu_{\text{x}}^{(n_0)} + \\
& + cov(x^{(n_0)}, m^{(n_0)})]^T + E[b^{(1)}, \dots, b^{(n0)}])
\end{align*}

Finally for elements $i,j$ with $j = 0, \dots, n_0$ and  $\rvec{w}^{(i)} \in \mathbb{R}^{n_0}$:
\begin{align*}
\sigma(E[&(\rvec{w}^{(i)})(\mu_m^{(j)} \mu_{\text{x}}^{(j)} + cov(x^{(j)}, m^{(j)})) + E[b^{(j)}])\\
&\geq \sigma(n_0\mu^{1}_{w}\mu_{m}\mu_{\text{x}} + \mu^{1}_{b}) + \sigma(n_0\mu^{1}_{w}cov(x^{(j)},m^{(j)}))
\end{align*}

Now let $T_1 = \sigma(n_0\mu^{1}_{w}Cov(\rvec{x}^{1},\rvec{m}^{1}))$, where $Cov(a,b)$ denotes the vector of element-wise covariances between items of $a$ and $b$ then:
$$E[\rvec{h}^1] \geq \sigma(n_0\mu^{1}_{w}\mu_{m}\mu_{\text{x}} + \mu^{1}_{b}) + T_1.$$

This term is propagated through all the network layers, adding bias to the model's output. (Full proof in the Appendix). A straightforward way to tackle the issue would be altering the model's input, by a subtraction debias: 
$$\rvec{x_{new}} = \rvec{x} - [cov(x^{(1)}, m^{(1)}),\dots, cov(x^{(n)}, m^{(n)})]^T$$

thus alleviating the MAR-derived effect ($T1$ term). Subsequently, one could apply sparsity normalisation, as proposed by \citet{vsp}. This way, the mean of the network's output is not explicitly dependent on the the missingness rate over all data instances. However, that term is intractable since it requires the full features $x^{(i)}$ to be known.
Instead, we propose a simple mechanism that takes into account both the input feature values and their missingness pattern, and evaluates the similarity between data instances with the same, or similar, structure.

\vspace{2mm}
\textbf{Core Idea:} \textit{Latent factors conditioned on the value and missingness vector, including the covariance between each input feature and the respective missingness indicator can be approximated through transformer-like similarity scores between non-linear embedding projections under the MAR assumption.}
\vspace{2mm}






Instead of the transformation :  $$\rvec{x_{new}} = (\rvec{x} - \rvec{\beta}) / \alpha $$
where $\rvec{\beta}$ is the per-feature covariance term and $\alpha$ the per-instance sparsity normalization scaling, we opt for an embedding mechanism.
The mechanism is conditioned on the fused value-existence input vectors $(\rvec{x} \odot \rvec{m}, \rvec{m})$, and creates a common representation embedding space $\textit{K} \subset \mathbb{R}^{emb}$ for the input instances.
Multihead self-attention is employed towards that goal, since, it can use the per-instance available context to retrieve the appropriate non-affine transformations from the embedding space ($\rvec{trans} \in \textit{K}$) and apply them to the provided input. In this way both the per-instance context (similar to $\alpha$), and the global context (similar to $\rvec{\beta}$) are incorporated in the final solution:
$$ lookup(\rvec{x}) = \rvec{trans} $$
$$\rvec{x_{new}} = transform(\rvec{x}, \rvec{trans})$$
where the lookup function, $\mathrm{lookup} : \mathbb{R}^{N} \to \mathbb{R}^{emb}$ is implemented with multihead self-attention, and the transform function, $\mathrm{transform} : \mathbb{R}^{(N+emb)} \to \mathbb{R}^{N}$ by non-linear feed-forward components.
\section{MAIN Algorithm}

The MAIN algorithms consists of two complementary steps: The Positional Encoded Vector (PEV) creation, and the multihead attention step with opacity gating. 
Let $\rvec{x} \in \mathbb{R}^{N}$ denote an input vector of $N$ features, and $x_1, x_2, \dots , x_N$ the scalar values representing the value of each specific feature in  $\rvec{x}$. PEV encoding augments the initial feature vector in order to:
\begin{enumerate}
    \item Differentiate between the case of a "filled value", in case of a missing feature, i.e zero in zeta normalization scenarios, and the existing equivalent of that feature value in a feature vector - zero because the feature in that specific instance is equal to the mean of the distribution that generates it. 
    \item Create an orthogonal basis of feature existence. From this perspective, the original feature vector is treated as a linear combination of inter-changeable vectors whose direction is denoted by the feature position and magnitude by their respective feature's value. 
\end{enumerate}

Here, for the specified input length $N$, the minimum number of bits required to describe all available positions, $bw$ is first calculated.
Then, for each training example $\rvec{x}$ the following components are calculated:
\begin{enumerate}[a)]
    \item A binary mask, indicating the existence of each feature in $\rvec{x}$.
    \item A binary positional encoding mask, marking the position of each feature in $\rvec{x}$.
\end{enumerate}

Finally, we concatenate a),b) feature-wise with the original input. That results to a 2D input matrix for each training example $\rvec{x}_{augmented} \in \mathbb{R}^{{N} \times {\log_{2}(N+1)}}$. In that way, we provide a simple way for the model to capture the similarity between data instances based on both their values and their missingness patterns.
The aforementioned PEV procedure is described in  Algorithm 1.

\begin{algorithm}
\caption{PEV creation}\label{alg:pev}
\begin{algorithmic}[1]
\Function{PevMaskGenerator}{$x$}
\State $bw\gets \lceil{\log_2 n}\rceil$
\For{$i \gets 1$ to $n$}
\State $bin_i \gets bin_{bw}(i)$ 
\State $m_i \gets$ 0 if x[i] = $\emptyset$ else 1 
\State $pev_i \gets [m_i|bin_i]$ 
\EndFor\label{euclidendwhile}
\State $pev \gets [pev_1, pev_2, ... pev_n]$
\State $m \gets   [m_1, m_2, ... , m_n]$
\State \textbf{return} $pev, m$
\EndFunction
\end{algorithmic}
\end{algorithm}

\begin{figure}[H]
\centering
\captionsetup{justification=centering}
\includegraphics[width=\linewidth]{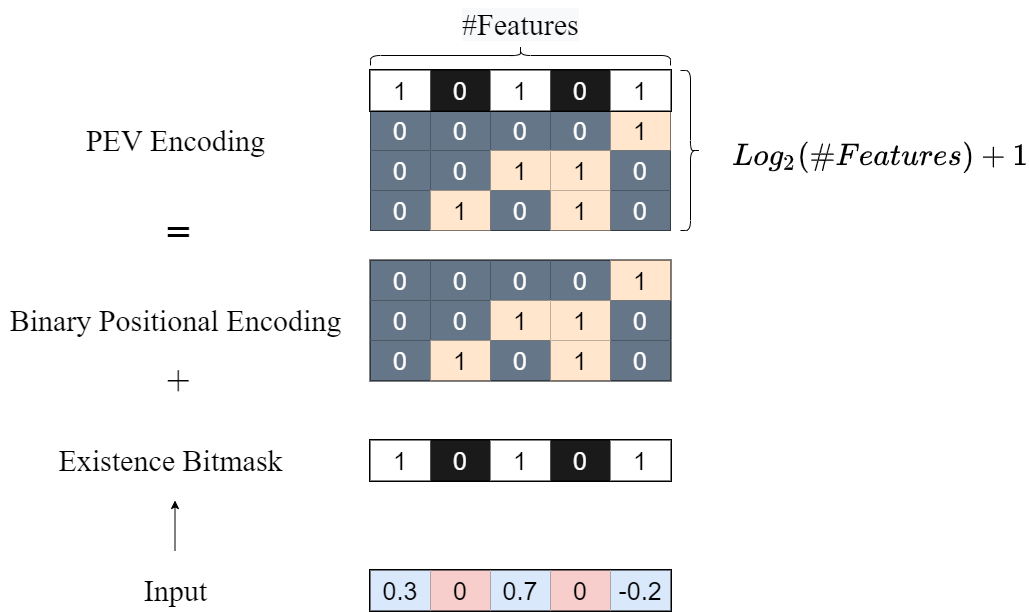}
\caption{Creation of PEV encoding Matrix.}
\label{fig:pev_creation}
\end{figure}

\begin{algorithm}
\caption{Opaque Multi Head Attention}\label{alg:main}
\begin{algorithmic}[1]
\While{$convergence$ $is$ $not$ $achieved$}
\For{$batch$ $in$ $X$}
    \For{$(x,y)$ $in$ $batch$}
        \State $pev, m \gets PevMaskGenerator(x)$
        \State $query \gets proj_n(dq([x:pev]))$ 
        \State $key  \gets proj_k(dk([x:pev]))$
        \State $value \gets proj_n(dv([x:pev]))$
        \State $imputed_{x} \gets mha(query,key,value)$
        \State $\hat{x} \gets \gamma \cdot imputed_{x} + (1-\gamma)\cdot query$ 
        \For {$dsm_j$ $in$ $Model$ $Tasks$}
            \State $\hat{y_j} \gets dsm_j(\hat{x})$
        \EndFor
        \State {$\hat{y} \gets [\hat{y_0}:\hat{y_1}:,\dots,:\hat{y_l}]$}
        \State {$loss \gets e(y,\hat{y})$}
        \EndFor
    \State {$batch$ $loss \gets \sum loss$}
    \State {$backprop(batch$ $loss)$}
\EndFor
\State \textbf{return}
\EndWhile
\end{algorithmic}
\end{algorithm}

\begin{figure}[H]
\includegraphics[width=\linewidth]{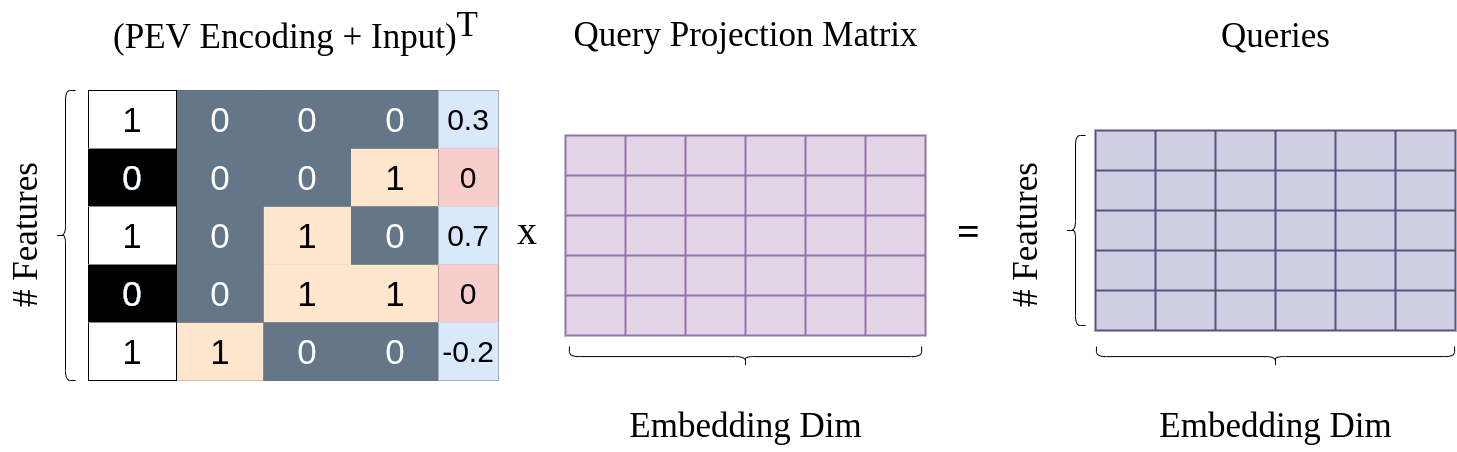}
\caption{Creation of Query Projections.}
\label{fig:query_creation}
\end{figure}

\begin{figure}[H]
\includegraphics[width=\linewidth]{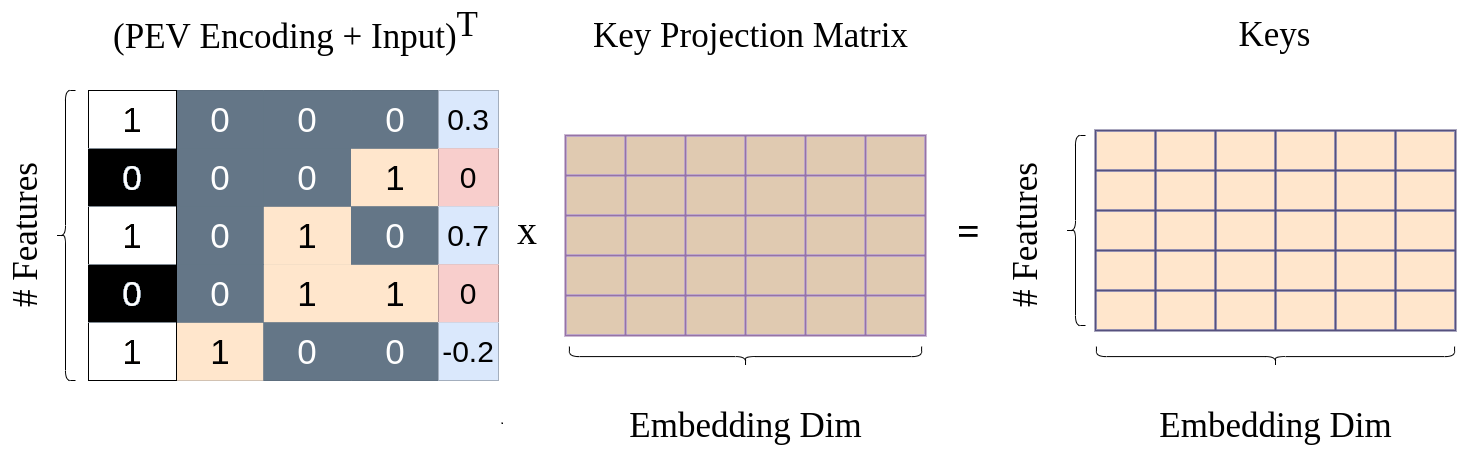}
\caption{Creation of Key Projections.}
\label{fig:key_creation}
\end{figure}

\begin{figure}[H]
\includegraphics[width=\linewidth]{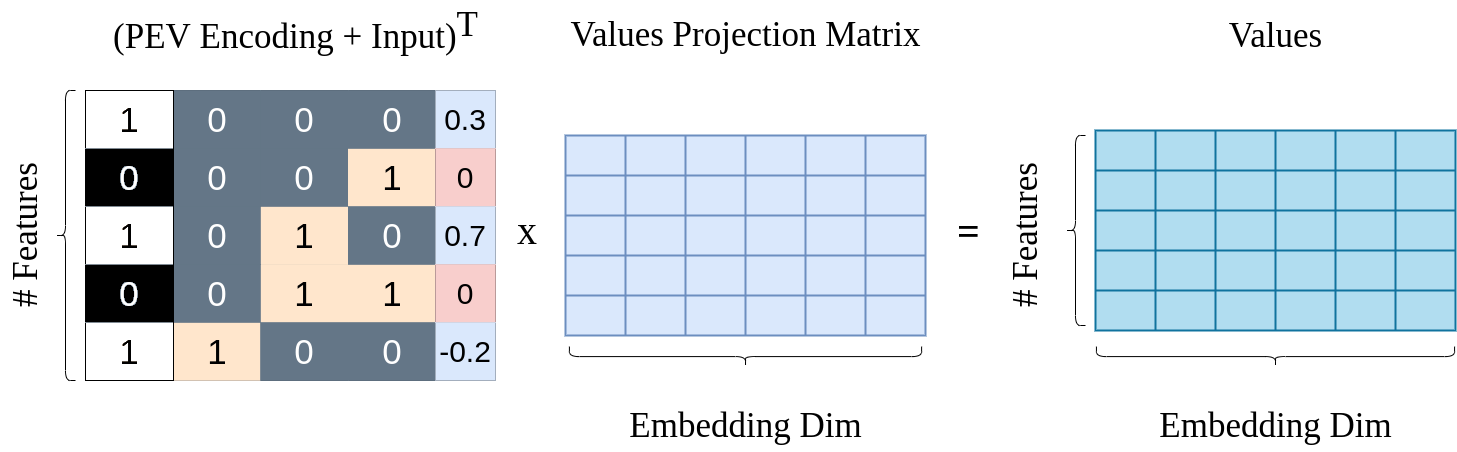}
\caption{Creation of Value Projections.}
\label{fig:value_creation}
\end{figure}

\begin{figure*}[t]
\centering
    \includegraphics[width=0.85\linewidth]{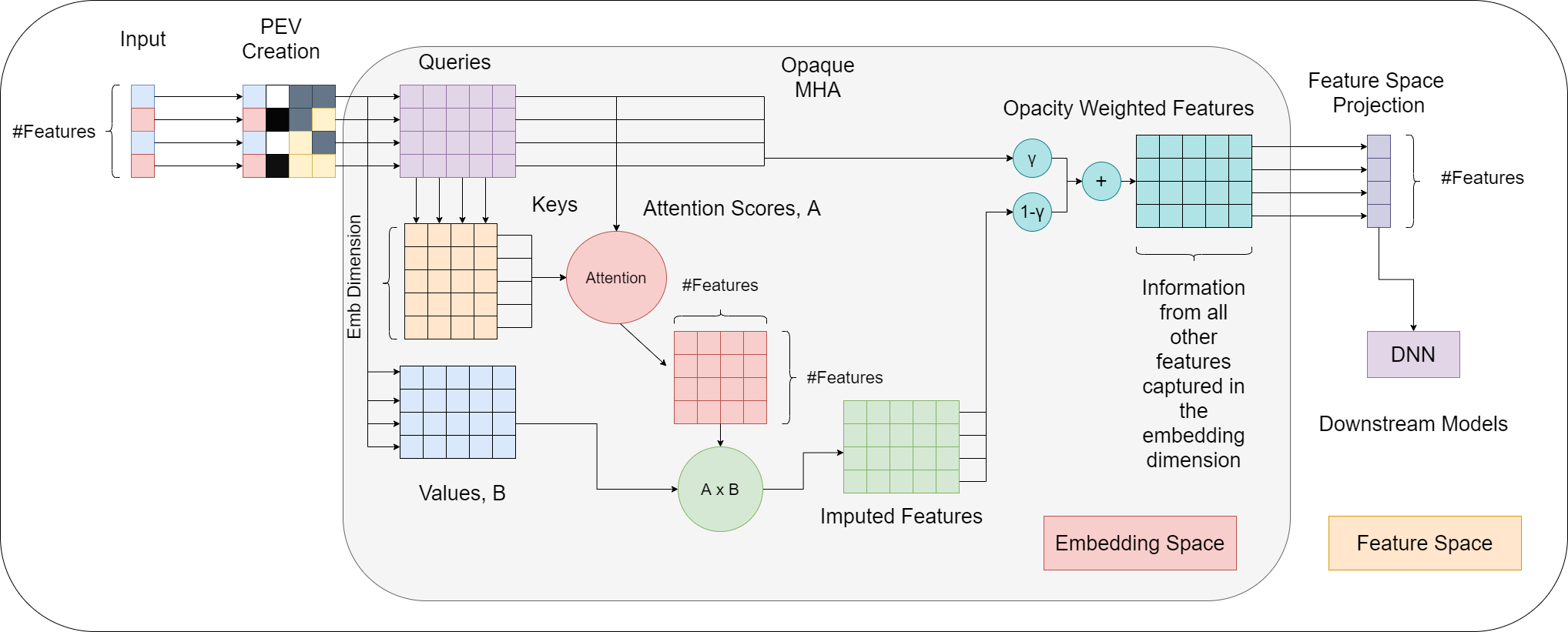}
    \caption{An overview of MAIN's architecture. It is comprised of three basic modules: a) PEV creation, b) Opaque Multi-head Attention, c) Downstream Tasks/Models}
    \label{fig:attention}
\end{figure*}

The next step is a Luong-style \cite{luong} attention mechanism of each individual augmented feature over the total augmented feature vector, effectively serving as self-attention. 
Note that the attention function is considered a mapping procedure between a Query and a set of Key-Value pairs \cite{aiayn}. In that fashion, we project the augmented input (PEV + features) into 3 embedding vector spaces, Query, Key and Value, using trainable projections. Here, the PEV augmentation will act as a mechanism capable to guide the focus of the self-attention heads into retrieving saliency from the existence-position prespective. 
The example of Query projections creation is depicted in Figure \ref{fig:pev_creation}, and the Keys and Values are created similarly (Figures \ref{fig:key_creation}, \ref{fig:query_creation}). That of course, could not be possible without the projection of all features to a common embedding space. This is enabled through the use of PEV embeddings, that transform each feature from a mere scalar to a querable vector.
This is the key link with the $T_N$ bound of Theorem 2. The correlation between each feature value and the fact that it is missing, is conditioned on the values of the other features in the original measurement. That is in the end composed out of the each feature's value, and whether it exists or not in the specific instance.

As a next step, we perform multi-head attention over each augmented feature, noted as \textbf{\textit{mha}} in Algorithm \ref{alg:main}, resulting in the imputed feature vector $imputed_x$.

Finally, we employ a trainable gating mechanism called opacity gate, $\gamma(\cdot)$. The gate acts as a "trainable sigmoid knob" between the original input of the model, and the imputed input produced by the MAIN mechanism.
The gate is unconditioned on the input features, and initialized at 0.5, giving equal importance to both inputs. During training, the gate is left free to decide through backprop whether our method provides an input more "salient" towards minimizing the downstream loss (Gate $ \simeq 0 $) or if the original input is better (Gate $ \simeq 1 $). Furthermore, the gate acts in a twofold manner, providing an abstraction buffer towards the next layers. During the first iterations, the next layers will try to adapt towards an unoptimal solution to the task, based on the original inputs while the MAIN mechanism is still in early training stage. This is not permanent however, since the gate will shift towards the MAIN inputs, "fine-tuning" the rest of the model into the optimal solution.
 
 The algorithm finishes with collecting all the opacity weighted feature vectors, $\hat{x}$ and feeding them to 
 an arbitrary number of downstream models, noted as \textbf{\textit{dsm}} in  Algorithm \ref{alg:main}. They are jointly trained with the MAIN component to perform the required predictive tasks, and during this analysis are all considered to be 1, since no-multitask dataset was used.

\section{Experiments}

Our method was evaluated on 7 Tabular datasets, 6 of which come from the UCI Dataset Repository and 1 maritime dataset related to vessel's performance which is property of DeepSea Technologies (i.e DeepSea V9). 
For all experiments, in order to be comparable with other methods that simulate missingness, binary sampling was performed per feature using the same technique as in \cite{vsp} and \cite{gain}. Our method though is not bound to this restriction and can work directly on missing data, exploiting the full capacity of the dataset. The downstream task in the UCI datasets was binary classification with imbalanced classes, while in the maritime dataset was regression. It is noteworthy to outline here that out of the 6 datasets, Breast, Credit, Spam and Heart are injected with MCAR missingness at 4 different levels: $[20\%,40\%,60\%,80\%]$, while Pima, Mammographic and DeepSea V9 have instrinsic MAR missingess at levels: $[12.2\%,3.35\%,7.28\%]$. In order for all the experiments to be compared at a common basis, the full rows of those datasets were polluted so that the total missing rate matched the $[20\%,40\%,60\%,80\%]$ scale, as a mixture of MCAR + MAR missingness.

It is important to outline that the scope of the experiments is towards to a \textbf{single-step} end-to-end training scheme. We found that the comparison with classical \textbf{two-step}, methods that first-impute then train/test formulations (i.e KNN, MICE, MCMC) is not only less relevant to the scope of this work but also unfair to them since they don't optimize for a specific task and give inferior results. For the sake of completeness however, we opt to compare with the \textbf{two-step tabular} method, GAIN \cite{gain}, since its authors shared our common interest of optimizing towards a downstream task. We also compared with SN \cite{vsp} as our method builds on the same ideas and can be seen as an extension to theirs. ZIMC (i.e zero imputation with mask concat) is serving as a baseline and FiLM \cite{fwt} although not an imputation method, since it also inspired our work can be seen as an intermediate step between SN and our work MAIN. We choose not to compare with AimNet \cite{aimnet}, as there was no official codebase, making it very difficult to repliate results. Furthermore,it was a DataBase-Oriented imputation method for the project HoloClean, not a general purpose Machine Learning Solution like the other compared methods.

The experimental design is similar to \cite{vsp} and \cite{aimnet}, reporting the test AUROC (UCI) and MSE (DeepSea V9) score of 5 runs. 
In order to keep the comparison fair, the GAIN \cite{gain} method was trained and tested on different splits of the dataset unlike the original implementation where training and evaluation was in the same set.
The Breast, Spam, Credit and DeepSea V9 datasets were normalized in the range of [0,1], while Heart, Pima were normalized in the range [-1,1] as in \cite{vsp}, \cite{gain}.

We used no explicit prepossessing for categorical values in the above datasets since our scheme: a) creates automatically trainable embeddings for both scalar and categorical/ordinal variables and b) we don't reconstruct the missing feature and thus we don't have to map reconstructed logits to corresponding classes as in \cite{gain}. 

Regarding train/validation/test splits, in most imputation-only approaches the RMSE reconstruction metric is usually reported on a single dataset without any splits. Since no official splits exist in any UCI datasets, instead of the 70-20-10 split used in \cite{vsp} we opt for a 70-30 split with 10-fold stratified cross-validation.

Models Setup:
In order to produce comparable results in UCI and test the performance gains of our imputation method as a base layer, we tried to keep the total number of layers (imputation + downstream) of our scheme close to what proposed in \cite{selu} as the most appropriate for the UCI and also used in \cite{vsp}; namely 4 Hidden layers @ 256 units and Adam Optimizer. The same principle applied to all the compared schemes where we had: imputation method + 4 layers dedicated to the downstream task.

Due to class imbalance, in our training scheme we pre-calculate the class weights of the target classes and use them into a weighted binary crossentropy loss, penalizing  misclassification on the minority class more heavily that misclassifications on the majority class. Finally, the reported metric is the AUROC curve interpolated at 200 points with Riemann summation method.

\
\begin{table}[h]
\renewcommand{\arraystretch}{1.4}
\label{tb_breast}
\fontsize{6}{7}\selectfont
\centering
\captionsetup{singlelinecheck=off}
\caption{: UCI Breast-Wisconsin}
\begin{tabular}{l@{\hskip6pt}c@{\hskip8pt}c@{\hskip8pt}c@{\hskip8pt}c}
\hline
\textbf{MR} & \textbf{20\%} & \textbf{40\%}  & \textbf{60\%} & \textbf{80\%}\\
\hline
ZIMC                   & 0.9501 $\pm$ 0.005 & 0.9485 $\pm$ 0.007 & 0.9182 $\pm$ 0.010 & 0.8470 $\pm$ 0.012 \\
GAIN$^{\star}$                    & 0.9872 $\pm$ 0.008  & 0.9475 $\pm$ 0.011 & 0.9171 $\pm$ 0.035 & 0.8443 $\pm$ 0.017 \\
SN      & 0.9683 $\pm$ 0.007 & 0.9341 $\pm$ 0.009 & 0.8593 $\pm$ 0.029 & 0.8640 $\pm$ 0.027 \\
FiLM   & 0.9818 $\pm$ 0.003 & 0.9513 $\pm$ 0.005 & 0.9263 $\pm$ 0.008 & 0.8757 $\pm$ 0.015 \\
\textbf{MAIN}                        & \textbf{0.9821 $\pm$ 0.003} & \textbf{0.9786 $\pm$ 0.005} & \textbf{0.9693 $\pm$ 0.013} & \textbf{0.9241 $\pm$ 0.014} \\
\hline
\end{tabular}
\end{table}

\begin{table}[!htb]
\renewcommand{\arraystretch}{1.4}
\caption{: UCI Credit Dataset}
\label{tb_credit}
\centering
\captionsetup{singlelinecheck=off}
\fontsize{6}{7}\selectfont
\begin{tabular}{l@{\hskip6pt}c@{\hskip8pt}c@{\hskip8pt}c@{\hskip8pt}c}
\hline
\textbf{MR} & \textbf{20\%} & \textbf{40\%}  & \textbf{60\%} & \textbf{80\%}\\
\hline
ZIMC        & 0.7297 $\pm$ 0.008 & 0.7051 $\pm$ 0.007 & 0.6833 $\pm$ 0.010 & 0.6349 $\pm$ 0.013 \\
GAIN$^{\star}$                      & 0.7412 $\pm$ 0.008 & 0.7173 $\pm$ 0.013 & 0.6849 $\pm$ 0.008 & 0.6019 $\pm$ 0.032 \\
SN       & 0.7396 $\pm$ 0.005 & 0.7145 $\pm$ 0.007 & 0.6826 $\pm$ 0.003 & 0.6332 $\pm$ 0.012 \\
FiLM   & 0.7443 $\pm$ 0.005 & 0.7187 $\pm$ 0.007 & 0.7025 $\pm$ 0.005 & 0.6528 $\pm$ 0.011 \\
\textbf{MAIN}                         & \textbf{0.7456 $\pm$ 0.004} & \textbf{0.7209 $\pm$ 0.006} & \textbf{0.7032 $\pm$ 0.004} & \textbf{0.6577 $\pm$ 0.009} \\
\hline
\end{tabular}
\end{table}

\begin{table}[!htb]
\renewcommand{\arraystretch}{1.4}
\centering
\captionsetup{singlelinecheck=off}
\caption{: UCI Spam Dataset}
\label{tb_spam}
\fontsize{6}{7}\selectfont
\begin{tabular}{l@{\hskip6pt}c@{\hskip8pt}c@{\hskip8pt}c@{\hskip8pt}c}
\hline
\textbf{MR} & \textbf{20\%} & \textbf{40\%}  & \textbf{60\%} & \textbf{80\%}\\
\hline
ZIMC                 & 0.9740 $\pm$ 0.005 & 0.9519 $\pm$ 0.007 & 0.9243 $\pm$ 0.008 & 0.8675 $\pm$ 0.011 \\
GAIN$^{\star}$                      & 0.9645 $\pm$ 0.004 & 0.9451 $\pm$ 0.008 & 0.9213 $\pm$ 0.001 & 0.8623 $\pm$ 0.010 \\
SN     & 0.9798 $\pm$ 0.002 & 0.9568 $\pm$ 0.002 & 0.9270 $\pm$ 0.002 & 0.8707 $\pm$ 0.005 \\
FiLM   & 0.9735 $\pm$ 0.005 & 0.9571 $\pm$ 0.006 & 0.9276 $\pm$ 0.007 & 0.8725 $\pm$ 0.012 \\
\textbf{MAIN}                        & \textbf{0.9817 $\pm$ 0.005} & \textbf{0.9589 $\pm$ 0.007} & \textbf{0.9298 $\pm$ 0.008} & \textbf{0.8732 $\pm$ 0.011} \\
\hline
\end{tabular}
\end{table}

\begin{table}[h]
\renewcommand{\arraystretch}{1.4}
\centering
\captionsetup{singlelinecheck=off}
\caption{\textsc{: UCI Pima Dataset}}
\label{tb_pima}
\fontsize{6}{7}\selectfont
\begin{tabular}{l@{\hskip6pt}c@{\hskip8pt}c@{\hskip8pt}c@{\hskip8pt}c}
\hline
\textbf{MR} & \textbf{20\%} & \textbf{40\%}  & \textbf{60\%} & \textbf{80\%}\\
\hline
ZIMC                  & 0.8131 $\pm$ 0.016 & 0.7864 $\pm$ 0.014 & 0.7585 $\pm$ 0.015 & 0.7001 $\pm$ 0.015 \\
GAIN$^{\star}$                       & 0.8074 $\pm$ 0.017 & 0.7861 $\pm$ 0.020 & 0.7503 $\pm$ 0.023 & 0.6987 $\pm$ 0.022 \\
SN     & 0.8121 $\pm$ 0.013 & 0.7851 $\pm$ 0.021 & 0.7589 $\pm$ 0.017 & 0.7006 $\pm$ 0.015 \\
FiLM  & 0.8017 $\pm$ 0.017 & 0.7725 $\pm$ 0.015 & 0.7475 $\pm$ 0.016 & 0.6913 $\pm$ 0.010 \\
\textbf{MAIN}                          & \textbf{0.8442 $\pm$ 0.009} & \textbf{0.7981 $\pm$ 0.007} & \textbf{0.7628 $\pm$ 0.011} & \textbf{0.7183 $\pm$ 0.012} \\
\hline
\end{tabular}
\end{table}

\begin{table}[!htb]
\renewcommand{\arraystretch}{1.4}
\centering
\captionsetup{singlelinecheck=off}
\caption{: UCI Heart Dataset}
\label{tb_heart}
\fontsize{6}{7}\selectfont
\begin{tabular}{l@{\hskip6pt}c@{\hskip8pt}c@{\hskip8pt}c@{\hskip8pt}c}
\hline
\textbf{MR} & \textbf{20\%} & \textbf{40\%}  & \textbf{60\%} & \textbf{80\%}\\
\hline
ZIMC                   & 0.8256 $\pm$ 0.009 & 0.8002 $\pm$ 0.012 & 0.6752 $\pm$ 0.028 & 0.6702 $\pm$ 0.033 \\
GAIN$^{\star}$                       & 0.8419 $\pm$ 0.039  & 0.7586 $\pm$ 0.065 & 0.6399 $\pm$ 0.102 & 0.6357 $\pm$ 0.094 \\
SN     & 0.8533 $\pm$ 0.007 & 0.7974 $\pm$ 0.012 & 0.7570 $\pm$ 0.007 & 0.6619 $\pm$ 0.014 \\
FiLM  & 0.8530 $\pm$ 0.007 & 0.7915 $\pm$ 0.014 & 0.7516 $\pm$ 0.009 & 0.6422 $\pm$ 0.017 \\
\textbf{MAIN}                         & \textbf{0.8702 $\pm$ 0.008} & \textbf{0.8371 $\pm$ 0.007} & \textbf{0.7851 $\pm$ 0.008} & \textbf{0.6802 $\pm$ 0.011} \\
\hline
\end{tabular}
\end{table}

\begin{table}[!htb]
\renewcommand{\arraystretch}{1.4}
\centering
\captionsetup{singlelinecheck=off}
\caption{: UCI Mammographic Dataset}
\label{tb_mamm}
\fontsize{6}{7}\selectfont
\begin{tabular}{l@{\hskip6pt}c@{\hskip8pt}c@{\hskip8pt}c@{\hskip8pt}c}
\hline
\textbf{MR} & \textbf{20\%} & \textbf{40\%}  & \textbf{60\%} & \textbf{80\%}\\
\hline
ZIMC                   & 0.7851 $\pm$ 0.014 & 0.7433 $\pm$ 0.027 & 0.6712 $\pm$ 0.026 & 0.6049 $\pm$ 0.026 \\
GAIN$^{\star}$                       & 0.8010 $\pm$ 0.024  & 0.7501 $\pm$ 0.036  & 0.7126 $\pm$ 0.028  & 0.6448 $\pm$ 0.057 \\
SN    & 0.7794 $\pm$ 0.011 & 0.7367 $\pm$ 0.021 & 0.6753 $\pm$ 0.014 & 0.6063 $\pm$ 0.019 \\
FiLM  & 0.8198 $\pm$ 0.012 & 0.7472 $\pm$ 0.023 & 0.6860 $\pm$ 0.042 & 0.6350 $\pm$ 0.032 \\
\textbf{MAIN}                         & \textbf{0.8807 $\pm$ 0.005} & \textbf{0.8535 $\pm$ 0.009} & \textbf{0.7998 $\pm$ 0.012} & \textbf{0.7682 $\pm$ 0.014} \\
\hline
\end{tabular}
\end{table}

\begin{table}[!htb]
\renewcommand{\arraystretch}{1.4}
\centering
\captionsetup{singlelinecheck=off}
\caption{: DeepSea V9}
\label{tb_ds9}
\fontsize{6}{7}\selectfont
\begin{tabular}{l@{\hskip6pt}c@{\hskip8pt}c@{\hskip8pt}c@{\hskip8pt}c}
\hline
\textbf{MR} & \textbf{20\%} & \textbf{40\%}  & \textbf{60\%} & \textbf{80\%}\\
\hline
ZIMC                  & 0.0013 $\pm$ 0.0005 & 0.0030 $\pm$ 0.0006 & 0.0124 $\pm$ 0.0010 & 0.0280 $\pm$ 0.0013\\
GAIN$^{\star}$                       & \textbf{0.0009 $\pm$ 0.0002} & \textbf{0.0027 $\pm$ 0.0002} & 0.0086 $\pm$ 0.0001 & 0.0267 $\pm$ 0.0004 \\
SN     & 0.0012 $\pm$ 0.0002 & 0.0048 $\pm$ 0.0003 & 0.0116 $\pm$ 0.0005 & 0.0310 $\pm$ 0.0009 \\
FiLM  & 0.0010 $\pm$ 0.0004 & 0.0035 $\pm$ 0.0005 & 0.0091 $\pm$ 0.0005 & 0.0270 $\pm$ 0.0009 \\
\textbf{MAIN}                           & 0.0009 $\pm$ 0.0003 & \textbf{0.0027 $\pm$ 0.0002} & \textbf{0.0085 $\pm$ 0.0002} & \textbf{0.0261 $\pm$ 0.0008} \\
\hline
\end{tabular}
\end{table}
\begin{figure*}[t]
\begin{center}
  \includegraphics[width=0.3\linewidth]{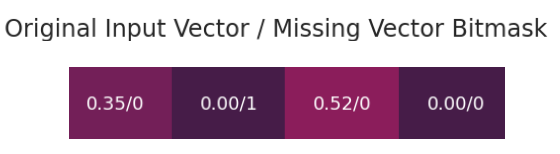}  
\end{center}
\caption{An illustrated example of a training example with missing features. (feat.value/missing)}
\label{fig:ablationinput}
\end{figure*}

\begin{figure*}[t]
\begin{center}
  \includegraphics[width=0.90\linewidth]{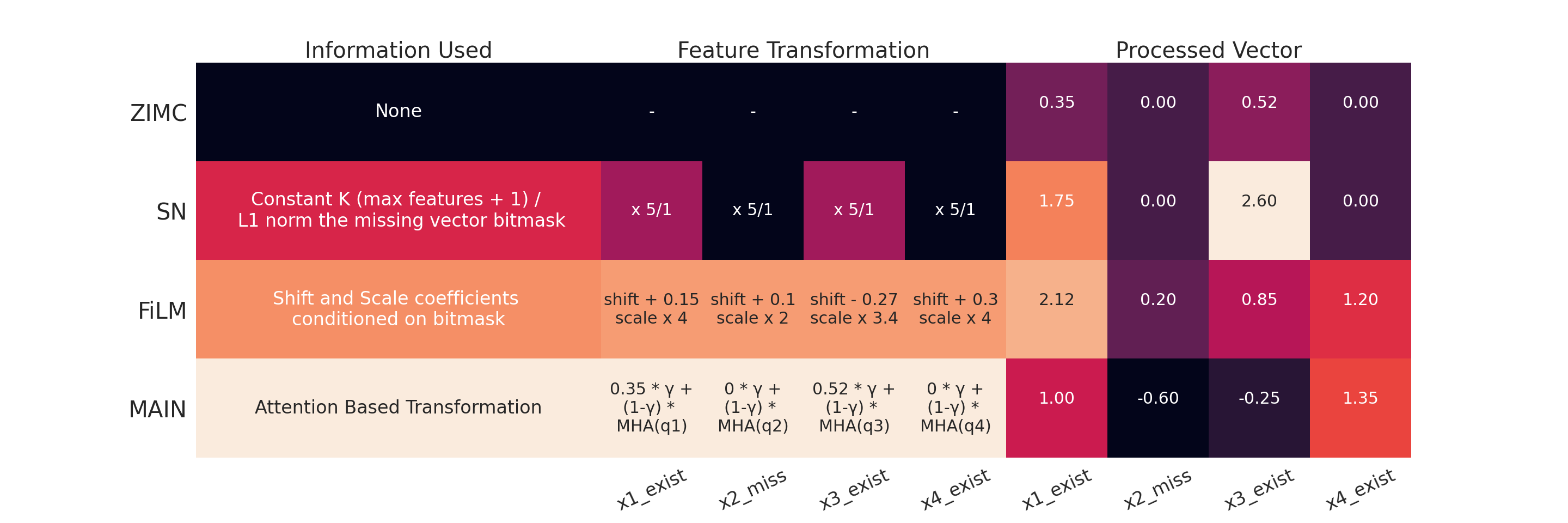}  
\end{center}
\caption{An illustrated example of the methods under test.}
\label{fig:ablation}
\end{figure*}
As Tables I-IV show, both GAIN and SN provide a far better alternative to the ZIMC baseline in every dataset and missing rate. While FiLM is of subpar performance towards both the generative GAIN and the discriminative SN in most cases, it is noteworthy, that in the case of the heavy MAR dataset (Pima), FiLM gives worse results to SN, while with none of the current state-of-the-art methods solving explicitly the MAR case.

On the contrary, the MAIN method seems to be vastly outperforming other methods, especially in the case of high missing rates, where it maintains considerably higher scores than its counterparts. In the case of the DeepSea V9 regression task, GAIN and MAIN have the least test MSE and perform equally in the low missing scenarios, while MAIN slightly outperforms GAIN in the high missing scenarios. 

\section{Discussion}
In this section we will give an illustrated example of a partially missing feature vector in order to provide more insights on how our method works compared to other well-known methods. The setup is the following: in figure \ref{fig:ablationinput}, we present a z-normalized 4D feature vector that will be provided as an input to the different methods that were compared against our final solution, MAIN.

The 4D input vector consists of 3 non-missing features: [$x1, x3, x4$] with corresponding values [$0.35, 0.52, 0.00$], and one missing feature: $x2$. The respective existence bitmask of the input is depicted with a binary indicator after the "slash" symbol ('/') of each feature, where '1' means missing feature, thus the corresponding mask is $m=[0,1,0,0]$.

In figure \ref{fig:ablation} we break-down how the input information is potentially transformed (i.e Feature Transformation) by 4 different approaches, leading to a the Processed Vector on the right in the same figure. The processed vector can be seen as the result of a learnable scaling factor (for the most algorithms) which forms the input to the following-up machine learning method (e.g downstream model). The actual numbers are figurative just to illustrate the underlying mechanics of each method. 

We observe that in the case of ZIMC as is the most naive method features are not explicitly affected since it is a mere zero imputation technique. It must be noted here, that the mask concatenation in ZIMC, potentially provides to the underline model with joint knowledge of missingness but that could happen in a deeper layer interactions in the case of neural nets. The even simpler zero-imputation couldn't discriminate at all, between a missing feature from a feature placed at the mode of the distribution in the z-scaled space. 

In SN, the inverse L1 norm of the bitmask scales all the non-zero features, by multiplying by their number + 1 (K) which is an improvement over zero-fill. On the other hand the scaling of existing features at the mode of the distribution (i.e zero-mean like $x4$ ) are affected in the same way as the missing ones (i.e $x2$), re-scaling only the rest of them in order to alleviate the VSP problem\cite{vsp}. In FiLM, since all the features are  subject to affine transformation conditioned in the missing context (feature + mask), the missing $x2$ and non-missing $x4$ zero-valued features are treated differently. Thus we claim that FiLM can be seen as a feature-wise rescaling extension of the SN which learns how to condition missing context to independent feature scalers.  Following this logic and going one step further, if we explicitly define the conditioning mechanism (i.e attention) we have MAIN which offers even more expressivity to learn the underline mechanics of the missing distribution.
 
\section{Conclusion}
In this work, we proposed MAIN a novel method based on multi-head attention to deal with missing data in continuous or discrete datasets. Our method works in a single step by implicitly imputing missing data and is optimized directly on the downstream task, offering an end-to-end trainable system. We demonstrate that MAIN significantly outperforms state-of-the-art methods in a variety of open dataset and also in a proprietary one.


\FloatBarrier
\bibliography{references.bib}

\begin{thebibliography}{16}
\providecommand{\natexlab}[1]{#1}
\providecommand{\url}[1]{\texttt{#1}}
\expandafter\ifx\csname urlstyle\endcsname\relax
  \providecommand{\doi}[1]{doi: #1}\else
  \providecommand{\doi}{doi: \begingroup \urlstyle{rm}\Url}\fi

\bibitem[Yi et~al.(2020)Yi, Lee, Kim, Hwang, and Yang]{vsp}
Joonyoung Yi, Juhyuk Lee, Kwang~Joon Kim, Sung~Ju Hwang, and Eunho Yang.
\newblock Why not to use zero imputation? correcting sparsity bias in training
  neural networks.
\newblock In \emph{7th International Conference on Learning Representations},
  2020.

\bibitem[Dumoulin et~al.(2018)Dumoulin, Perez, Schucher, Strub, de~Vries,
  Courville, and Bengio]{fwt}
Vincent Dumoulin, Ethan Perez, Nathan Schucher, Florian Strub, Harm de~Vries,
  Aaron Courville, and Yoshua Bengio.
\newblock Feature-wise transformations.
\newblock \emph{Distill}, 2018.

\bibitem[Wu et~al.(2020)Wu, Zhang, Ilyas, and Rekatsinas]{aimnet}
Richard Wu, Aoqian Zhang, Ihab Ilyas, and Theodoros Rekatsinas.
\newblock Attention-based learning for missing data imputation in holoclean.
\newblock 2020.

\bibitem[Little and Rubin.(2014)]{missingness}
Roderick~JA Little and Donald~B Rubin.
\newblock \emph{Statistical analysis with missing data}.
\newblock John Wiley \& Sons, 2014.

\bibitem[Li et~al.(2019)Li, Jiang, and Marlin]{misgan}
Steven~Cheng{-}Xian Li, Bo~Jiang, and Benjamin~M. Marlin.
\newblock Misgan: Learning from incomplete data with generative adversarial
  networks.
\newblock In \emph{7th International Conference on Learning Representations},
  2019.

\bibitem[Mattei and Frellsen(2019)]{MIWAE}
Pierre-Alexandre Mattei and Jes Frellsen.
\newblock {MIWAE}: Deep generative modelling and imputation of incomplete data
  sets.
\newblock In \emph{Proceedings of the 36th International Conference on Machine
  Learning}, Proceedings of Machine Learning Research, pages 4413--4423, Long
  Beach, California, USA, 09--15 Jun 2019. PMLR.

\bibitem[Naz{\'{a}}bal et~al.(2018)Naz{\'{a}}bal, Olmos, Ghahramani, and
  Valera]{HI-VAE}
Alfredo Naz{\'{a}}bal, Pablo~M. Olmos, Zoubin Ghahramani, and Isabel Valera.
\newblock Handling incomplete heterogeneous data using vaes.
\newblock \emph{CoRR}, 2018.

\bibitem[Pathak et~al.(2016)Pathak, Kr{\"a}henb{\"u}hl, Donahue, Darrell, and
  Efros]{inpaint}
Deepak Pathak, Philipp Kr{\"a}henb{\"u}hl, Jeff Donahue, Trevor Darrell, and
  Alexei~A. Efros.
\newblock Context encoders: Feature learning by inpainting.
\newblock \emph{2016 IEEE Conference on Computer Vision and Pattern Recognition
  (CVPR)}, 2016.

\bibitem[Bora et~al.(2018)Bora, Price, and Dimakis]{ambient}
Ashish Bora, E.~Price, and A.~Dimakis.
\newblock Ambientgan: Generative models from lossy measurements.
\newblock In \emph{ICLR}, 2018.

\bibitem[Ghiasi et~al.(2018)Ghiasi, Lin, and Le]{dropblock}
Golnaz Ghiasi, Tsung-Yi Lin, and Quoc~V Le.
\newblock Dropblock: A regularization method for convolutional networks.
\newblock In \emph{Advances in Neural Information Processing Systems 31}, pages
  10727--10737. 2018.

\bibitem[Yoon et~al.(2018)Yoon, Jordon, and van~der Schaar]{gain}
Jinsung Yoon, James Jordon, and Mihaela van~der Schaar.
\newblock Gain: Missing data imputation using generative adversarial nets.
\newblock In \emph{6th International Conference on Learning Representations},
  2018.

\bibitem[He et~al.(2015)He, Zhang, Ren, and Sun]{he}
Kaiming He, Xiangyu Zhang, Shaoqing Ren, and Jian Sun.
\newblock Delving deep into rectifiers: Surpassing human-level performance on
  imagenet classification.
\newblock In \emph{Proceedings of the 2015 IEEE International Conference on
  Computer Vision (ICCV)}, 2015.

\bibitem[\'{S}mieja et~al.(2018)\'{S}mieja, Struski, Tabor, Zieli\'{n}ski, and
  Spurek]{bengio}
Marek \'{S}mieja, \L~ukasz Struski, Jacek Tabor, Bartosz Zieli\'{n}ski, and
  Przemys\l~aw Spurek.
\newblock Processing of missing data by neural networks.
\newblock In \emph{Advances in Neural Information Processing Systems 31}, pages
  2719--2729. 2018.

\bibitem[Luong et~al.(2015)Luong, Pham, and Manning]{luong}
Thang Luong, Hieu Pham, and Christopher~D. Manning.
\newblock Effective approaches to attention-based neural machine translation.
\newblock In \emph{Proceedings of the 2015 Conference on Empirical Methods in
  Natural Language Processing}, Lisbon, Portugal, 2015.

\bibitem[Vaswani et~al.(2017)Vaswani, Shazeer, Parmar, Uszkoreit, Jones, Gomez,
  Kaiser, and Polosukhin]{aiayn}
Ashish Vaswani, Noam Shazeer, Niki Parmar, Jakob Uszkoreit, Llion Jones,
  Aidan~N Gomez, \L~ukasz Kaiser, and Illia Polosukhin.
\newblock Attention is all you need.
\newblock In \emph{Advances in Neural Information Processing Systems 30}. 2017.

\bibitem[Klambauer et~al.(2017)Klambauer, Unterthiner, Mayr, and
  Hochreiter]{selu}
G{\"{u}}nter Klambauer, Thomas Unterthiner, Andreas Mayr, and Sepp Hochreiter.
\newblock Self-normalizing neural networks.
\newblock In \emph{Advances in Neural Information Processing Systems 30: Annual
  Conference on Neural Information Processing Systems 2017, Long Beach, CA,
  {USA}}, 2017.

\end{thebibliography}

\end{document}